\documentclass[runningheads]{llncs}
\usepackage{comment}
\usepackage{amsmath,amssymb} % define this before the line numbering.
\usepackage{color}
\usepackage{dsfont}
\usepackage{wrapfig}
\usepackage{graphicx}
\usepackage{subcaption}
\DeclareMathOperator*{\argmax}{argmax}
\DeclareMathOperator*{\argmin}{argmin}

\begin{document}

\pagestyle{headings}
\mainmatter

\title{Improving Feature Attribution through Input-specific Network Pruning} 
%
% \begin{comment}
\titlerunning{ }
% If the paper title is too long for the running head, you can set
% an abbreviated paper title here
%
\author{Ashkan Khakzar\inst{1} \and
 Soroosh Baselizadeh\inst{1} \and
 Saurabh Khanduja\inst{1} \and
 Christian Rupprecht\inst{3} \and
 Seong Tae Kim\inst{1} \and
 Nassir Navab\inst{1,2}}
% %
\authorrunning{ }
\institute{Technical University of Munich, Germany \and
Johns Hopkins University, USA \and
University of Oxford, UK
}
% \end{comment}
%******************
\maketitle
%%%%%%%%% ABSTRACT%%%%%%%%%%%%%%%%%%%%%%%%%%%%%%%%%%%%%%%%%%5
\begin{abstract}
Attributing the output of a neural network to the contribution of given input elements is a way of shedding light on the black-box nature of neural networks. 
Due to the complexity of current network architectures, current gradient-based attribution methods provide very noisy or coarse results.
We propose to prune a neural network for a given single input to keep only neurons that highly contribute to the prediction.
We show that by input-specific pruning, network gradients change from reflecting local (noisy) importance information to global importance. 
Our proposed method is efficient and generates fine-grained attribution maps. 
We further provide a theoretical justification of the pruning approach relating it to perturbations and validate it through a novel experimental setup.
Our method is evaluated by multiple benchmarks: sanity checks, pixel perturbation, and Remove-and-Retrain (ROAR). 
These benchmarks evaluate the method from different perspectives and our method performs better than other methods across all evaluations.

\end{abstract}
%%%%%%%%%%%%%%%%%%%%%%%%%%%%%%%%%%%%%%%%%%%%%%%%%%%%%%%%%%
\section{Introduction}
With steadily improving task performance, understanding the prediction process of neural networks is becoming more and more important. The task of \emph{attribution} is to determine which part or region of the input is the most responsible for the prediction. 
The intended purpose for attribution methods is predominantly to measure \emph{global importance}~\cite{srinivas2019full}. The importance of an element is measured relatively --- by comparing the output with and without the element being present. Intuitively, an element is more important if its absence changes the output drastically.
The main difficulty is the combinatorial explosion of element groups that need to be removed. Consider the example of image classification. Removing a single pixel at a time is not enough since the object in the image can still be easily identified if only one pixel is missing. This means that in theory, one needs to remove all possible combinations of elements to be able to compute the region with the highest global importance, which is of course infeasible. 

There are two major branches in attribution literature on how to deal with this computational complexity. Gradient-based attribution~\cite{bach2015pixel,shrikumar2017learning,sundararajan2017axiomatic,montavon2017explaining,ancona2017towards} computes global importance by using model gradients as a proxy.
However, these estimation methods tend to be very noisy, especially with very large networks.

The second branch is based on input/image perturbation where pixels are removed/greyed out and the actual change in prediction is measured. The main research effort here is to efficiently test plausible groups of pixels, often by enforcing priors such as smoothness of the perturbation masks ~\cite{fong2017interpretable,fong2019understanding,qi2019visualizing,petsiuk2018rise,wagner2019interpretable}. 

In this paper, we extend the removal of features to hidden representations, by finding and keeping the neurons that their removal has the highest effect on the prediction. The hidden neurons correspond to multiple input elements and capture their interdependence. 
Instead of pruning the network for the whole dataset, we efficiently prune the network for a single prediction.
% by using neuron gradient magnitudes as a proxy for neuron importance. 
The pruning requires a single backward pass through the network and is thus very fast to compute. We observe that by pruning, the input gradients start reflecting global importance information, hence global features are uncovered from the noisy attribution maps that initially represent local features. We show this phenomenon both visually in attribution maps and quantitatively by evaluating the importance of features.
%
%
%%%%%--------------------------
\begin{figure}[t]
\centering
% \begin{center}
% \fbox{\rule{0pt}{2in} \rule{0.9\linewidth}{0pt}}
   \includegraphics[width=1\linewidth]{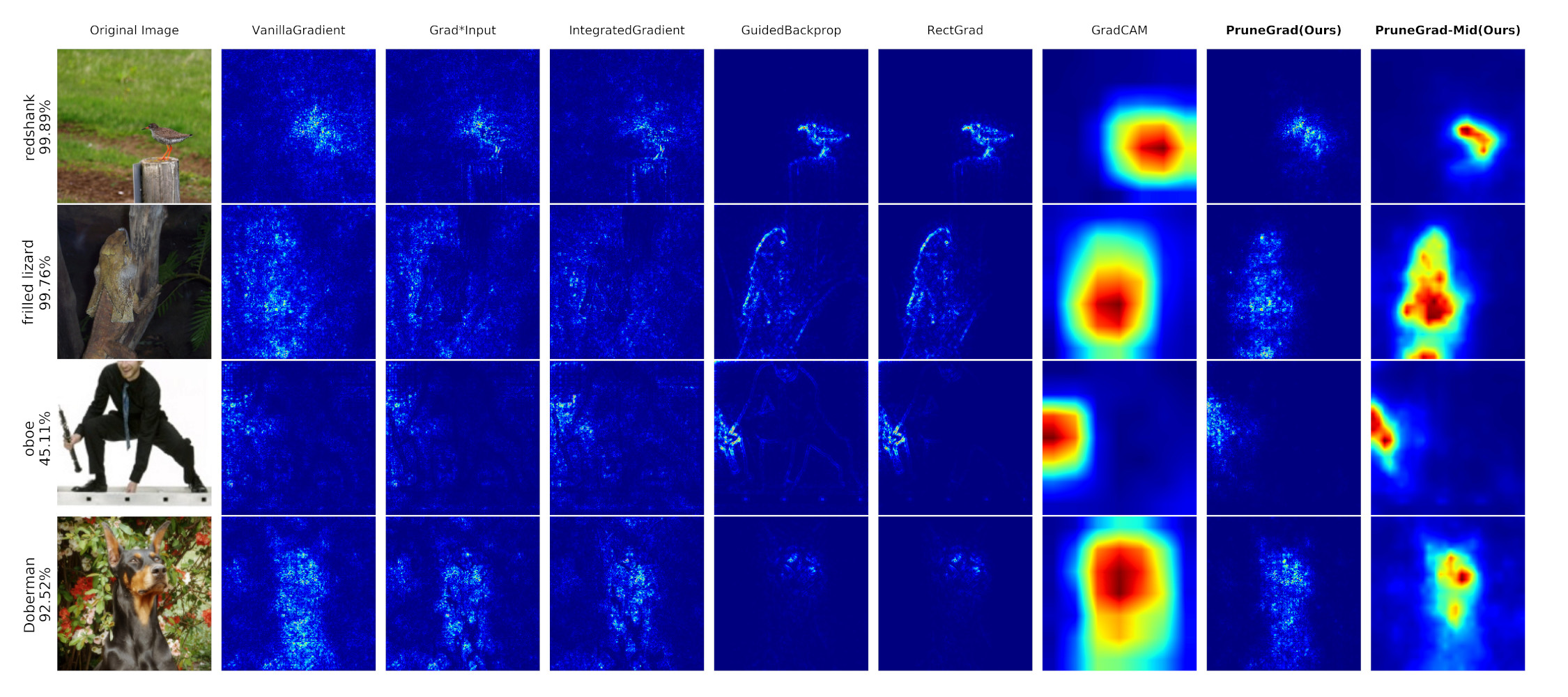}
% \end{center}
   \caption{\textbf{Visual evaluation} of our methods against gradient-based methods on ResNet-50 (ImageNet). The visual evaluation is not reliable, as some methods provide results that are more \emph{human}-interpretable however they fail quantitative evaluations. It is crucial to quantitatively evaluate whether the method is reflecting model behavior and whether the highlighted features are in fact contributing features for the model's prediction (Sec.~\ref{experiments}). For further examples on ResNet-50 and VGG-16 please refer to the supplementary materials.} 
\label{fig:qual}
\end{figure}
%%%%%--------------------------
% 
We provide a mathematical analysis relating our method to adversarial perturbations, which delivers more insights on how pruning helps to improve attribution. We posit that the gradient of the pruned network is a sufficiently close solution for the aforementioned adversarial perturbations. In order to experimentally validate the hypothesis, we use the recently introduced feature importance evaluation metric (Remove-and-Retrain). This work is the first to evaluate the adversarial perturbations using such feature importance evaluation metric, and explore the link between pruning, interpretation, and adversarial examples. Therefore our paper provides a new perspective for understanding networks and adversarial examples.

Evaluating attribution methods has considerably improved in the last years. Since the quality of visual explanations is subjective and hard to measure, a suite of quantitative benchmarks and sanity checks has been developed to assess a different facet of attribution methods. Adebayo et al.~\cite{adebayo2018sanity} introduce sanity checks that examine attribution maps before and after randomizing network weights. If the attribution map does not change considerably when using random weights, the attribution is not explaining model behavior. Passing these sanity checks is a necessary requirement for attribution methods.
Further, pixel perturbation~\cite{samek2016evaluating} and Remove-and-Retrain (ROAR)~\cite{roar_neurips,hooker2018evaluating} evaluate whether the highlighted features by the method are in fact highly contributing features for the network. We report pixel perturbation results on ImageNet, BirdSnap and CIFAR-10 datasets, and ROAR on BirdSnap and CIFAR-10. 
To our knowledge, this paper is the first to report results on all these benchmarks quantitatively and give a comprehensive overview. Our method achieves top-performance in all benchmarks.

\section{Related Work}
\subsection{Evaluation of Attribution Methods}  
% We start by introducing the evaluation and benchmarking tools for evaluating attribution methods to emphasize their importance. 
Early evaluations rely on human perception of what is interpretable. 
However, there is the caveat that although the attributions seem reasonable to humans, they do not necessarily reflect model behavior. 
Nie et al.~\cite{nie2018theoretical} showed theoretically and experimentally that certain attribution methods (GuidedBackProp~\cite{springenberg2014striving}, Deconvolution~\cite{zeiler2014visualizing}) with human interpretable attributions actually perform partial input recovery. 
\cite{adebayo2018sanity} further investigated this issue and laid out a set of sanity checks for attribution methods to pass. 
The sanity checks are experiments that evaluate the method's sensitivity to the model's parameter randomization and to label randomization. 
Several works \cite{sundararajan2017axiomatic,shrikumar2017learning,lundberg2017unified} propose evaluating the attribution methods by using theoretical axioms that are desirable for attribution methods to satisfy. 
Another group of evaluation methods adopt the notion of referenced-based importance directly in their evaluation. 
In a pioneering work, Samek et al.~\cite{samek2016evaluating} proposed removing pixels in the image based on the scores in the attribution map, and the effect on output shows whether the computed scores are reliable. 
As this effect on output might be as a result of the network not having seen the perturbed input during training, 
Hooker et al.~\cite{roar_neurips,hooker2018evaluating} further improved on this idea by introducing the Remove-and-Retrain (ROAR) framework, where the network is retrained on the modified inputs and the drop in accuracy is regarded as the effectiveness of the attribution method.    

\subsection{Gradient-Based Attribution Methods}  
% Although the reasoning and motivation behind these methods differ significantly, this category of methods leverage the gradient of output with respect to the input.
% \\
\textbf{Local Importance.} Simonyan et al.~\cite{simonyan2013deep} and Baehrens et a.~\cite{baehrens2010explain} assume a locally linear behavior of the model and propose the input gradient itself as a means of showing the importance of each input element 
%and the resulting saliency maps represent the local sensitivity of the model.  

\textbf{Modified Backpropagation Conditions.} GuidedBackProp~\cite{springenberg2014striving} and RectGrad~\cite{kim2019saliency} set specific conditions for backpropagating gradients. GuidedBackProp only allows positive gradients to be propagated at each neuron. It is shown that GBP performs partial image recovery and does not pass sanity checks\cite{adebayo2018sanity,nie2018theoretical}. 
RectGrad sets a more strict condition than GBP, and only allows backpropagating gradients when the product of that gradient and its corresponding activation are larger than a threshold.   

\textbf{Reference-based:} Another way to look at feature importance is to see the effect of not just the local change of input elements, but by changing to a reference value such as zero (i.e. removing that element). 
% This captures a more global perspective of importance and is not susceptible to effects such as feature saturation
% (When the gradient with respect to the feature is zero, however, the feature is making a contribution globally)
LRP~\cite{bach2015pixel} and DeepLift~\cite{shrikumar2017learning} use modified gradient propagation approaches for backpropagating the difference between output and reference output. Integrated-Gradients~\cite{sundararajan2017axiomatic} computes the contribution of each element, by integrating the gradients with respect to that element, while the element changes from reference to the current input.

\textbf{Using High-level Features:} These methods leverage hidden neurons and their gradients, hence capture high-level representations. GradCAM~\cite{selvaraju2017grad} and CAM~\cite{zhou2016learning} perform a weighted sum of last convolutional feature maps. Fullgrad~\cite{srinivas2019full} uses hidden biases and their corresponding gradients at each layer. 
%%%
\subsection{Perturbation-based Attribution Methods}
\textbf{Single/patch Occlusion.} These methods set one or multiple elements to a reference value. Zeiler et al.~\cite{zeiler2014visualizing} occlude a patch of pixels and observe the output change. Using a patch of pixels, as it captures the notion of multiple pixels as a feature, yields better results than single-pixel occlusion~\cite{ancona2017towards}. Observing output change by removing one single element does not take the interdependence between elements into account. One solution for this is using Shapely values method to find the contribution of each element. Due to the complexity of finding this solution, several works have proposed approximate solutions~\cite{lundberg2017unified,ancona2019explaining}. 
%However these methods approximate Shapley values for single elements, and not for sets of elements.  

\textbf{Mask Perturbation.} These methods mask the input with a certain reference value and aim at finding the smallest mask that keeps the output constant. 
%RISE~\cite{petsiuk2018rise} uses a number of random masks and these masks are summed weighted by their effect on the output change.
Fong et al.~\cite{fong2017interpretable,fong2019understanding} propose finding a mask that maximizes the output and regularizing the optimization with the size and smoothness of the mask. The smoothness prior avoids irregularly shaped masks. Qi et al. \cite{qi2019visualizing} improve the optimization process of finding the perturbation mask of ~\cite{fong2017interpretable} by using integrated gradients. Wagner et al.~\cite{wagner2019interpretable} propose an adversarial defense so that the optimization~\cite{fong2017interpretable} avoids adversarial perturbations. Fong et al.~\cite{fong2019understanding} further improve~\cite{fong2017interpretable} by changing the regularization terms into constraints. Schulz et al.~\cite{schulz2020restricting} perturb intermediate feature maps and quantify (in bits) the information in image regions.
%-------------------------------------------------------------------------
%
\section{Method}\label{method}
%% problem setup
\subsubsection{Problem Description.}
We study the problem of attributing the output of a neural network to the contribution of each input element. 
Formally, for a function $f: X\in \mathbb{R}^{n}\rightarrow \mathbb{R}$ that represents the output of neuron given input $X$, each element $a_{j}$ of an attribution map $A_{f}(X) \in \mathbb{R}^{n}$ is the contribution of input element $x_{j}$ to the output of $f$.

The principal idea of our method is to obtain the attribution from a simplified version of the network. 
This simplified network $f_p$ is obtained via pruning as detailed in the next section. 
%
%% ---------------------------------------------------------------
\subsection{Pruning for Feature Attribution}\label{method:pruning}
The main application for pruning neural networks is to increase inference speed and to decrease their memory footprint. 
Here, we use pruning for an input to simplify the neural network, such that the attribution does not get distracted by superfluous neurons.
To this end, we prune the network specifically for the given input $X$, as we are interested in discarding neurons that do not influence the current prediction.
This differs from the standard pruning procedure, where the network is pruned while taking the whole dataset into account. 

We formulate the problem of pruning a network for an input as a constrained optimization problem. 
As pruning removes neurons of $f$, we adapt the notation to $f_p(X, M)$, where $M \in \{0, 1\}^m$ effectively masks neurons to be removed during the computation of $f$. 
Thus, $m$ is the number of neurons inside the network and $f(X) = f_p(X, \mathds{1})$ and $f_p(X, 0) = 0$.

The task is now to find an $M$ that removes as many neurons as possible without strongly affecting the outcome of $f_p(X)$. 
Formally, the constrained optimization problem for input-specific pruning is then defined as:
\begin{equation}\label{eq:constrained_opt}
\begin{split}
\min_M |f_{p}(X, M) - f(X)| \\
s.t. \quad \left \| M \right \|_{0} \leq \kappa, \, M \in \left \{ 0,1 \right \}^{m}
\end{split}
\end{equation}
where $\kappa$ defines the sparsity level or the number of ones in $M$. 
We aim at finding an efficient, approximate solution for this constrained optimization problem. 
Therefore we propose a solution based on pruning of least contributing neurons. 
Similar to \cite{montavon2017explaining,figurnov2016perforatedcnns} we make two approximations.
\begin{equation}\label{eq:approx1}
    M^* = \argmin_M |f_{p}(X, M) - f(X)| \approx \argmin_M \sum_{i=1}^m M_i|f_{p}(X, e_i) - f(X)|
\end{equation}
In this first simplification, we assume that the mask can be found by analyzing neurons individually. 
Here, $e_j \in \{0, 1\}^m$ is an inverse indicator mask for deactivating only neuron $j$. 
That is $e_{j,i} = 1, \, \forall i \neq j$ and $e_{j,j} = 0$.
This allows us to reason about neuron contribution individually, reducing the otherwise exponential complexity of finding $M$ to $\mathcal{O}(m)$. 

The second approximation uses a first order Taylor expansion on $f_p$, substituted into \eqref{eq:approx1}.
\begin{equation}\label{eq:approx2}
    |f_{p}(X, e_i) - f(X)| \approx \left| f(X) - \frac{\partial f}{\partial n_i} n_i - f(X) \right| = \left|\frac{\partial f}{\partial n_i} n_i \right|
\end{equation}
Here, $n_i$ denotes the activation of neuron $i$. Thus, $\frac{\partial f}{\partial n_i}$ is the gradient of $f$ with respect to $n_i$ and can be computed with as single backward pass. 
Given a $\kappa$, one can then minimize \eqref{eq:approx1} by sorting the neurons by the scores obtained from \eqref{eq:approx2} and pruning the lowest $\kappa$ of them.

In our experiments, we find that the gradient $\nabla_{\delta}f_{p}$ of the pruned network $f_p(X, M^*)$ with respect to the input $X$ is much better suited for attribution than the gradient on the original network $f(X)$. 
In the next section, we provide an intuition on the reason for this improvement. 
Henceforth, we refer to this pruning and gradient strategy as PruneGrad.

\subsection{Connection to Input Perturbation}\label{method:reasoning}
We will draw a link to input perturbation methods, providing another angle to understand the effect of pruning. 
Each hidden neuron corresponds to a group of input elements and represents a feature (pattern) in the input. 
Perturbations in the input propagate through the network until they reach the target neuron. 
This means that changes in the input will affect many hidden neurons and form pathways of changes between input and target.
The pruned network is solely comprised of \emph{important} hidden neurons.
Therefore, in the pruned network, an input perturbation can only perturb the target neuron via perturbing important input features.   
Based on this intuition, in order to find important features in the input, we search for an input perturbation that maximizes the target neuron's output in the pruned network:    
\begin{equation}\label{eq:maximization}
\argmax_\delta f_p \left (X + \delta, M \right )
\end{equation}  
where $\delta$ is the perturbation, $\| \delta  \|_2 < \Delta$, and $\Delta$ is the upper bound for perturbation.
Linearizing $f_{p}$ through its gradient, $\nabla_{\delta}f_{p}$ is an approximate solution to Eq.~\ref{eq:maximization}.
In our experiments, we show that for the purpose of feature attribution $\nabla_{\delta}f_{p}$ works just as well as more accurate solutions.

%%%%% 
In the taxonomy of attribution based methods, PruneGrad is classified as a reference-based, gradient-based method. 
Though the perturbation of the input is performed locally by using the gradient, the contributions of hidden neurons are assigned relative to a baseline with removed hidden neurons. 
Therefore, the local input perturbations computed for the pruned model do not reflect the local sensitivity of the original model.    
%%%%%%
%%%%%%

Input perturbations on the original unpruned network are prone to exhibiting adversarial effects~\cite{goodfellow2014explaining,fong2017interpretable}. 
Wagner et al.~\cite{wagner2019interpretable} propose a defense against adversarial examples by limiting the neurons to the ones that were originally activated by the input. 
This is equivalent to pruning all neurons with zero activation in Eq.~\ref{eq:approx2}. 
The threshold that we use for pruning in all our experiments are higher than removing only zero contributing neurons.
We provide experiments in Sec.~\ref{exp:adversarial} to show the effect of input-specific pruning on adversarial perturbations.
%
%%%%%%%------------------------------------------------------------------------------
\subsection{Extension to Intermediate Layers}\label{method:pruneGradMid}
In previous sections, we showed how PruneGrad provides attribution maps for the input $X$. 
The same procedure can be extended to intermediate layers and activations (neurons) to determine their saliency scores. 
Let $\Lambda^{l} \in \mathbb{R}^{k_{l}\times h_{l} \times w_{l}} $ be the activations of the network at layer $l$, and $\Lambda_{p}^{l}$ be the corresponding activations in the pruned network. The corresponding attribution map for $\Lambda^{l}$ is:

\begin{equation}\label{eq:pruneGradCAM}
A_{f}(\Lambda^{l}) = \frac{\partial f_{p} }{\partial \Lambda_{p}^{l}}
\end{equation}

In the next section, we will evaluate PruneGrad for attribution qualitatively and quantitatively and analyze its hyper-parameters.
\section{Experiments and Results}\label{experiments}
The recently introduced benchmarks, sanity checks~\cite{adebayo2018sanity} and ROAR~\cite{roar_neurips} raised questions about the sanity and effectiveness of many attribution methods.
Each benchmark examines the methods from a different perspective. 
Sanity checks test whether the explanation map changes after randomizing the network parameters. 
If the map does not change, the attribution method is not explaining model behavior since randomized networks do not solve the task and thus show a different attribution.
Several previous attribution methods do not pass the sanity checks. 
To evaluate how much the highlighted features of attribution methods really contribute to the model's output, pixel perturbation and the ROAR benchmarks are used. 
In our setting, the former represents how well the method avoids attributing high scores to unimportant features, while the ROAR benchmark examines whether the highlighted features are important features over the entire dataset for the model to learn. 
%%-----------------------------------------------------------------------------------------------
\textbf{Comparisons.}
We compare our method with other, gradient based attribution methods: GradCAM~\cite{selvaraju2017grad}, the recently proposed RectGrad~\cite{kim2019saliency}, Integrated Gradients~\cite{sundararajan2017axiomatic}, GuidedBackProp (GBP)~\cite{springenberg2014striving}, Input$\times$Gradient~\cite{shrikumar2017learning} and pure gradient~\cite{simonyan2013deep} (VanillaGradient). 
GuidedGradCAM~\cite{selvaraju2017grad} is simply the multiplication of GradCAM and GBP and suffers from the same issues as GBP~\cite{adebayo2018sanity,nie2018theoretical,shrikumar2017learning}. 

We do not explicitly compare to LRP's z rule \cite{bach2015pixel} as it is equivalent to Input$\times$Gradient (up a scaling factor)~\cite{shrikumar2017learning,kindermans2016investigating}.
Similarly, DeepLift~\cite{shrikumar2017learning} is a fast approximation to Integrated Gradients~\cite{ancona2017towards}.  
%%%%%%%%%%%%%%%%
\subsection{Implementation Details}\label{implement-details}
\textbf{Pruning.} 
As detailed in Sec.~\ref{method:pruning}, we remove the neurons with the lowest importance scores. 
For a CNN, by ``neuron'' we refer to individual activations not entire activation maps (channels). 
We find the pruning threshold using a validation set. 
For ResNet-50, 1000 images from the ImageNet validation set are used to determine the threshold. 
For BirdSnap and CIFAR we use 10\% of training set.
The pruning threshold is chosen such that on average the effect on output change is equal to the allowed threshold $\kappa$.
Unless otherwise stated, we set output change threshold to 15\% (resulting in $\kappa=70\%$) for ResNet-50 on ImageNet. 
\textbf{Efficiency.} 
The deciding factor for the efficiency of an attribution method is the number of passes through the network.
VanillaGradient, Input$\times$Gradient, RectGrad, and GBP require one backpropagation step. 
IntegratedGradients requires 20 $\sim$ 50 steps \cite{sundararajan2017axiomatic}. 
Perturbation based methods, in contrast, require solving and optimization problem and therefore need significantly more backpropagation steps.
PruneGrad requires only two backpropagation steps but achieves better performance than other gradient-based methods.
\subsection{Pruning Helps Uncovering Global Features}\label{exp:pruning_level}
%%%%%-------------------------- output change vs sparsity
\begin{figure}[t]
% \fbox{\rule{0pt}{2in} \rule{0.9\linewidth}{0pt}}
\centering  
\begin{subfigure}[h]{0.35\linewidth}
     \includegraphics[width=\linewidth]{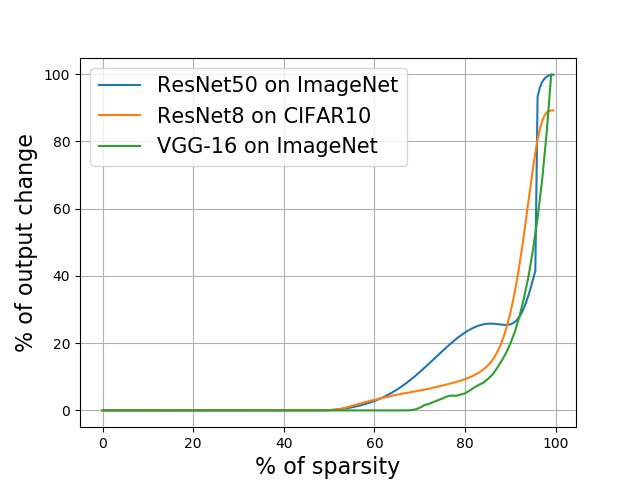}
     \caption{}
     \label{fig:sparsity_vs_output}
\end{subfigure}
\begin{subfigure}[h]{0.45\linewidth}
     \includegraphics[width=\linewidth]{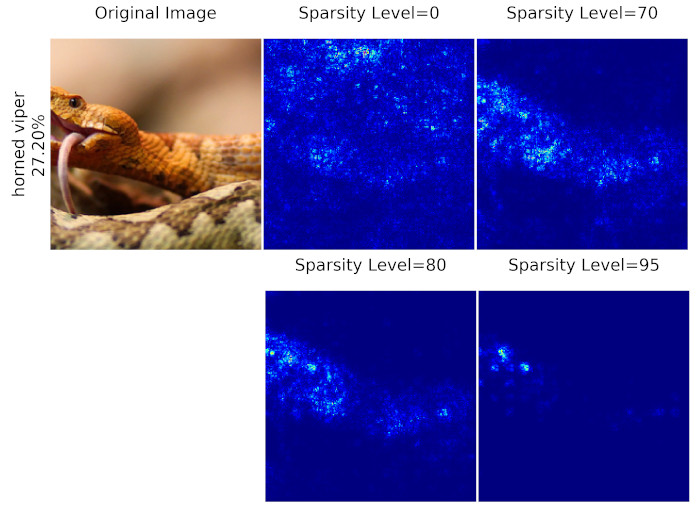}
     \caption{}
     \label{fig:pruning_threshold_qual}
\end{subfigure}
   \caption{Effect of pruning level (sparsity) on (a) the magnitude of output change, and (b) the generated attribution maps.}
\end{figure}
%%%%%--------------------------
We investigate the effect of the level of pruning $\kappa$ on the attribution maps and on the loss value in Eq.~\ref{eq:constrained_opt}. 
The loss value is a proxy for the magnitude of output change, and for more intuitive understanding we directly report the output change. 
Fig.~\ref{fig:sparsity_vs_output} shows the relationship between sparsity and output change for ResNet-50~\cite{he2016deep}, VGG-16\cite{simonyan2014very}, and ResNet-8. 
For ResNet-50 and VGG-16, the results are averaged over 1k images from the ImageNet validation set, and ResNet-8 results are the average over all CIFAR-10 validation images. 
Note that roughly 50\% of neurons can be removed without any change in output. 
The effect of pruning on the attribution maps shows interesting behavior, both visually and quantitatively. 
Fig.~\ref{fig:pruning_threshold_qual} shows this effect qualitatively. 
Without pruning ($\kappa=0$) the gradient reflects the local sensitivity of the model to the input change and highlights many areas in the background. 
When pruning, less relevant neurons are removed, and the gradient reflects more global importance of input features. 
Note that excessive pruning ($\kappa > 80\%$) removes highly contributing features and results in a significant output change (Fig.~\ref{fig:sparsity_vs_output}). 

Visually it is not possible to determine which threshold reflects the global features used by the network in the most complete way. 
Hence, we will use the ROAR framework in Sec.~\ref{exp:roar} to measure the importance of the detected input regions. 
The results of different pruning levels on the importance of reflected features are shown in Fig.~\ref{fig:prune_grad_pgd_diff_thresholds}.
Without pruning, which is equivalent to showing the local sensitivity of the model, the reflected features have minimal importance for the model. 
As the pruning increases, only important features are reflected, however excessive pruning again results in losing information in the attribution maps and inferior ROAR results (PruneGrad\_95 in Fig.~\ref{fig:prune_grad_pgd_diff_thresholds}).
\subsection{Visual Analysis}
Visual comparison to other gradient-based attribution methods is found in Fig.~\ref{fig:qual}. 
VanillaGradient reveals local sensitivity. 
The maps of IntegratedGradients and Input$\times$Gradient look similar to a combination of VanillaGradient and the input image. 
Possibly due to the dominant input term in their formulation. 

GuidedBackProp and RectGrad (which is a strict variant of GuidedBackProp) tend to visualize features that are very similar to the input images. 
Nie et al.~\cite{nie2018theoretical} proved that GuidedBackProp reconstructs low-level image features (e.g. edges) and is insensitive to model parameter randomization. 
It is shown in our sanity checks (Sec.~\ref{exp:sanity}) that these methods are insensitive to randomization and ROAR (Sec.~\ref{exp:roar}) shows their features are not important for the model.

In contrast, PruneGrad highlights fine-grained feature maps and it is visually evident that they correlate well with the objects of predicted classes.
In Sec.~\ref{exp:roar}, ROAR shows that these features are in fact important features for the model.
%
%
%-------------------------------------
\subsection{Sanity Checks}\label{exp:sanity}
%%%%%-------------------------- Sanity Checks
\begin{figure}[ht]
\centering
  \begin{subfigure}[h]{0.40\linewidth}
        \includegraphics[width=\linewidth]{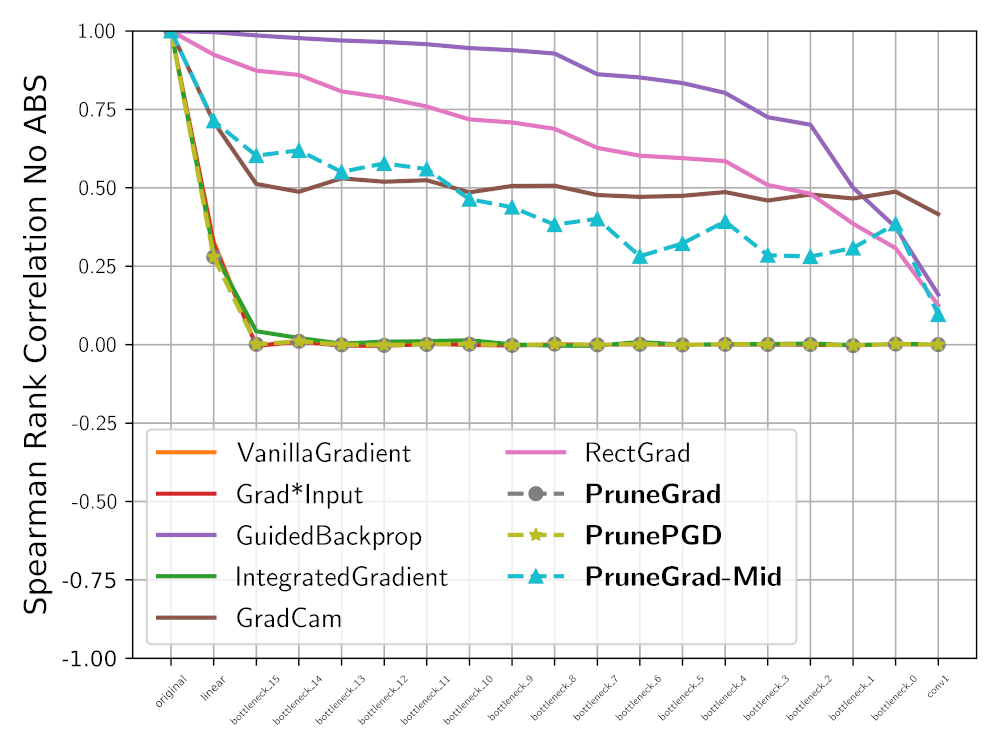}
        \caption{}
        \label{fig:a_plot}
  \end{subfigure}
%   \hfill
  \begin{subfigure}[h]{0.40\linewidth}
  \includegraphics[width=\linewidth]{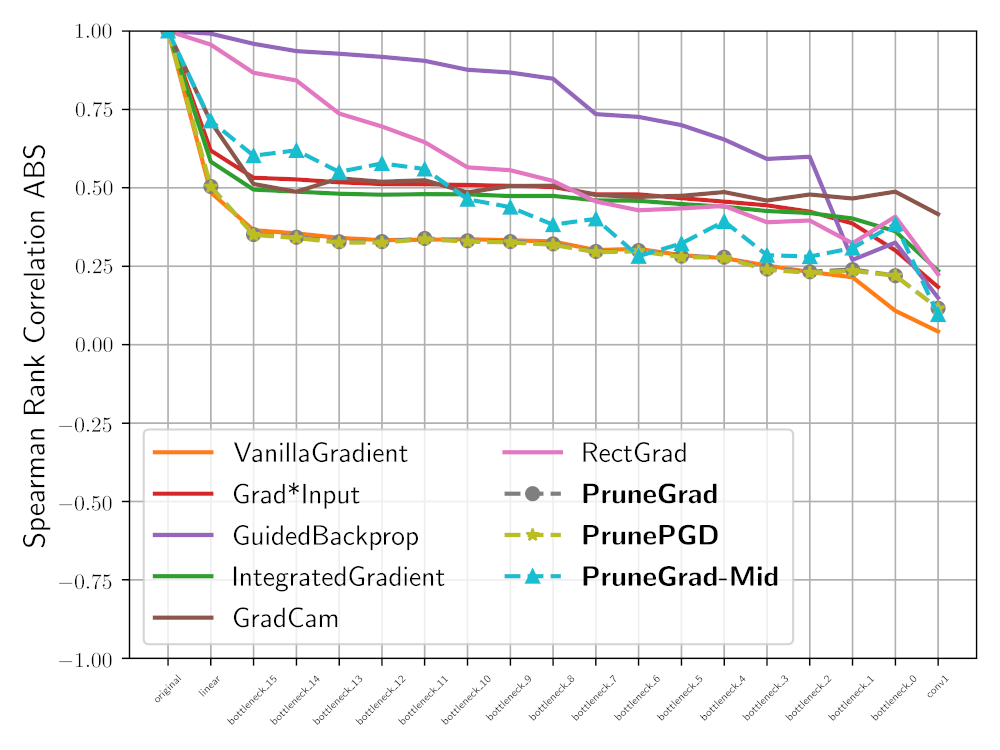}
        \caption{}
        \label{fig:b_plot}
  \end{subfigure}
%   \hfill
%   \begin{subfigure}[h]{0.30\linewidth}
%   \includegraphics[width=\linewidth]{images/SSIM.png}
%         \caption{}
%         \label{fig:c_plot}
%   \end{subfigure}
  \begin{subfigure}[h]{\linewidth}
   \includegraphics[width=1\linewidth]{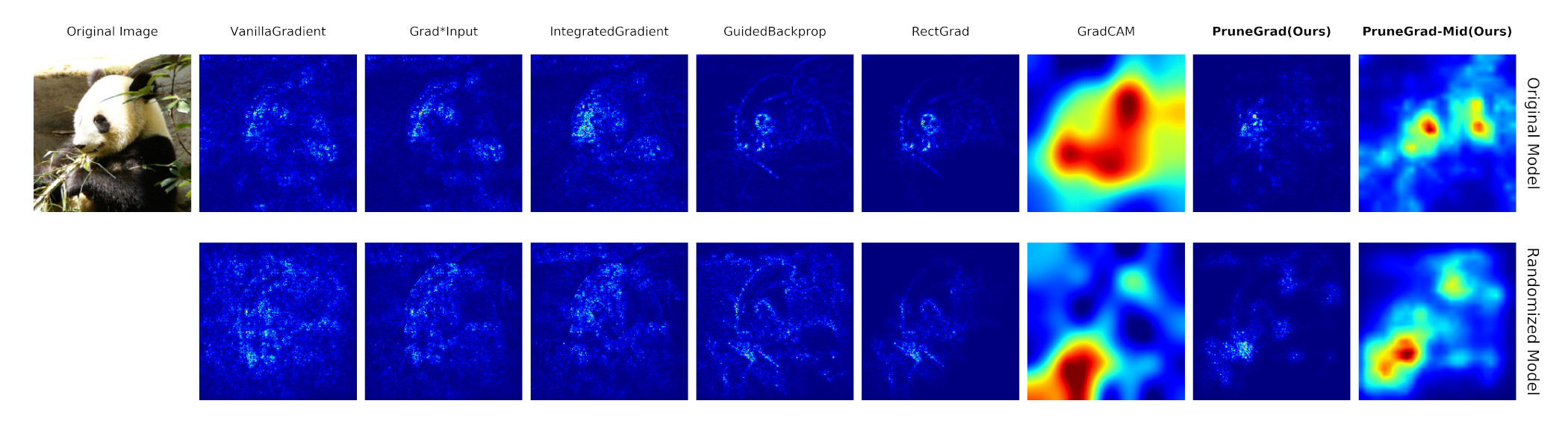}
   \caption{Attribution maps for an example image before and after full model randomization}
   \label{fig:sanity_qual}
\end{subfigure}
  \caption{\textbf{Sanity checks (ImageNet):}
  (c) The similarity between attribution maps before and after randomization implies that the attribution method is not explaining model behavior. (a), (b) Average results from 1k images from ImageNet on two similarity metrics. The x-axis shows up to which layer that randomization has been applied to on ResNet-50. Lower is better.}
  \label{fig:sanity}
\end{figure}
%%%%%--------------------------
%purpose
We conduct quantitative sanity checks~\cite{adebayo2018sanity} to evaluate the sensitivity of our methods to network parameter randomization. 
In this experiment, all learnable parameters of the network are successively replaced by random weights, from the last layer to the first layer (Fig. \ref{fig:a_plot}, \ref{fig:b_plot}). 
At each randomization step, the similarity between the attribution map generated from the original network is compared with the one obtained from the new randomized network. 
It is necessary that attribution methods are sensitive to such randomizations. 
We use a ResNet-50~\cite{he2016deep} network that is pre-trained on ImageNet~\cite{ILSVRC15} and reinitialize its parameters randomly with a normal distribution with zero mean and a standard deviation of 0.01. 
A random subset of 1k images from the ImageNet~\cite{ILSVRC15} test set is used.

The visual example in Fig. \ref{fig:sanity_qual} is provided for a better understanding of this experiment. 
The first row shows the original maps and the second row shows the attribution maps for the randomized models.
A good attribution map should show distinct differences in the two cases. 
%What's happening
In order to quantitatively compare the similarity of attribution maps (Fig. \ref{fig:a_plot}, \ref{fig:b_plot}) before and after parameter randomization, we use Spearman rank correlation (with and without applying an absolute function on the attribution maps). 
Lower similarity value indicates better performance in this test.
% emergency add-
% We normalize the attribution maps to range $[-1, 1]$ before calculating similarity scores in order to ignore the special characteristics of some methods as stated in ~\cite{adebayo2018sanity}. 
%-

%%%%disuccsing results
%% prunegrad and vanilla
Fig. \ref{fig:a_plot}, \ref{fig:b_plot} confirm the findings of Adebayo et. al~\cite{adebayo2018sanity} that VanillaGradients are the most sensitive.
PruneGrad performs equally well and better than all other methods. 
Additionally, VanillaGradient while being sensitive does not highlight important features and provides very noisy information (See ROAR and pixel perturbation experiments). 
The interpolation in GradCAM and PruneGrad-Mid (Sec.~\ref{exp:intermediate_layer}) results in coarse maps and hence the maps are structurally more similar before and after randomization. 
GradCAM and PruneGrad-Mid both suffer from this resolution issue in sanity checks.

%% rectgrad and GBP
Fig.~\ref{fig:sanity} provides further evidence that GuidedBackProp is not sensitive to model randomization~\cite{adebayo2018sanity,nie2018theoretical}. 
We also find that RectGrad which is a strict variant of GuidedBackProp also behaves similarly to GuidedBackProp and fails this model randomization test.
%% integrad, GradCAM
%
%
%
%
%
\subsection{Pixel Perturbation Benchmark}
%%%%%--------------------------Pixel Perturb
\begin{figure}[t]
% \fbox{\rule{0pt}{2in} \rule{0.9\linewidth}{0pt}}
\centering  
\begin{subfigure}[h]{0.3\linewidth}
     \includegraphics[width=\linewidth]{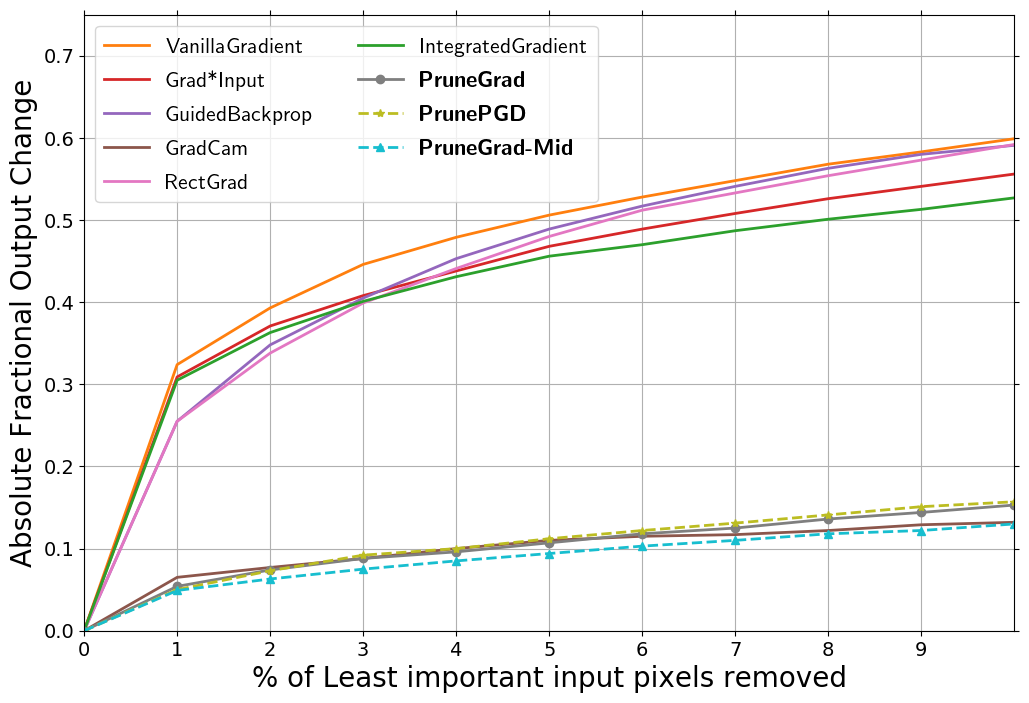}
     \caption{ImageNet}
     \label{fig:perturb_ImageNet}
\end{subfigure}
\begin{subfigure}[h]{0.3\linewidth}
     \includegraphics[width=\linewidth]{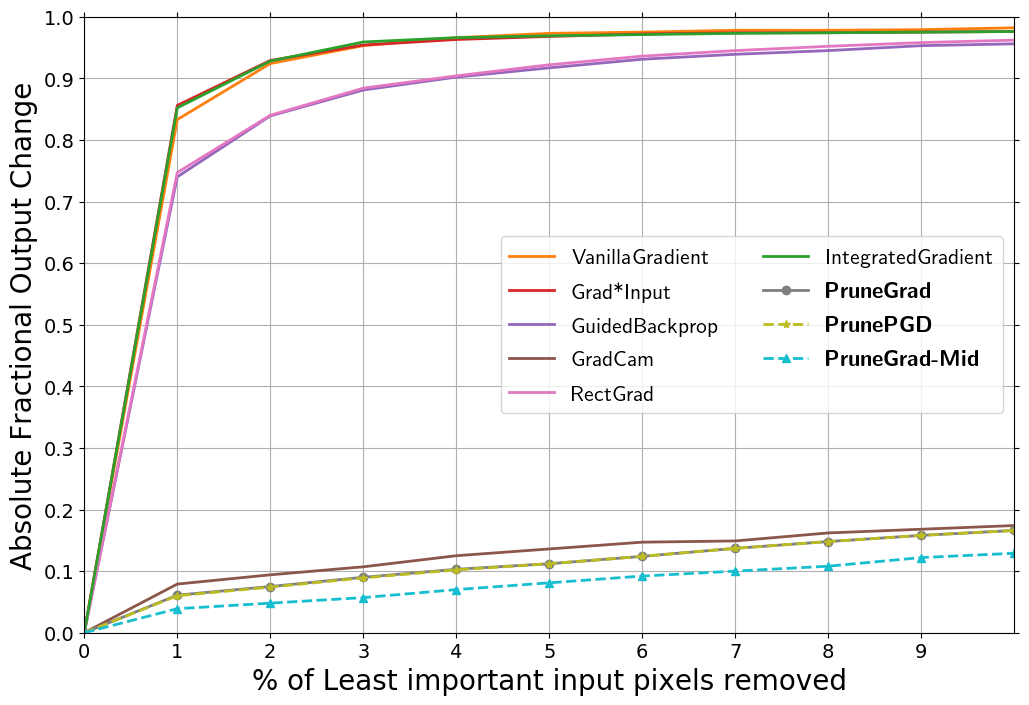}
     \caption{BirdSnap}
     \label{fig:perturb_BirdSnap}
\end{subfigure}
\begin{subfigure}[h]{0.3\linewidth}
     \includegraphics[width=\linewidth]{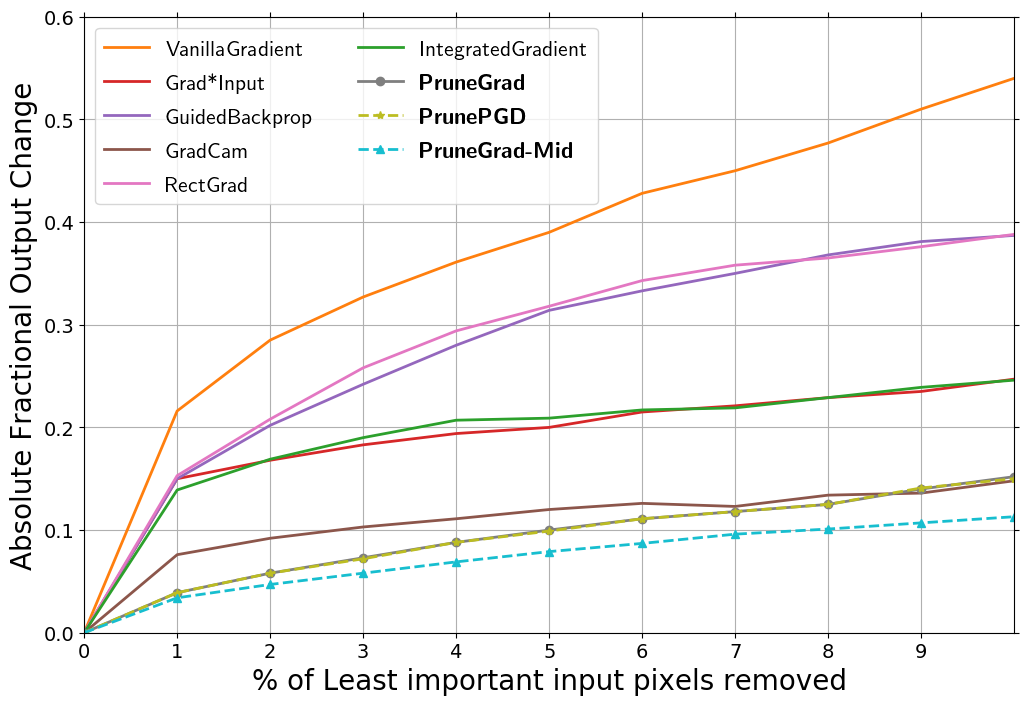}
     \caption{CIFAR-10}
     \label{fig:perturb_cifar}
\end{subfigure}
   \caption{\textbf{Pixel perturbation benchmark:} The effect of removing the least important pixels in the image (as determined by the attribution methods) on the absolute output change. Pixel perturbation is done on the original model for PruneGrad (similar to all). Lower is better.}
\label{fig:Perturb}
\end{figure}
%%%%%--------------------------
This experiment evaluates attribution methods by observing the effect of removing pixels (based on the scores provided by the methods) on the output of the model and is originally proposed by Samek et al. ~\cite{samek2016evaluating}. Srinivas et al. \cite{srinivas2019full} show that removing the pixels starting from the highest scores in descending order is more prone to producing artifacts for the network, therefore the output change is more likely to be a result of these artifacts than reflecting the importance of pixels. This claim is further supported in their experiments by showing that random attribution score assignment performs similar to other attribution methods if the pixels are removed in descending order. 
% This leads to inability of distinguishing a method which provides reasonable attributions with one that creates unnecessary artifacts.
Therefore in this section, we opt for removing pixels in ascending order, i.e. removing least important pixels first. 
In this way, the experiment reveals the specificity of different methods i.e.~if the low scoring elements reported by attribution methods are really unimportant. 
Note that, for our method, we perform the pruning only for finding attribution maps, after that the pixel perturbation is performed on the original model. 

In Fig.~\ref{fig:Perturb} we report results on CIFAR using a custom ResNet8 (three residual blocks), Birdsnap~\cite{berg-birdsnap-cvpr2014} using ResNet-50, and ImageNet (validation set) using ResNet-50. 
It shows the absolute fractional change of the output as we remove the least important pixels. 
Lower curves mean higher specificity of the methods. 
PruneGrad performs best in estimating unimportant pixels. 
This fact agrees with the intuition behind the pruning step of our framework in which we discard the unimportant features that contribute the least to the output.

%%%%---------------------
\subsection{Remove-and-Retrain (ROAR) Benchmark}\label{exp:roar}
%%%%%--------------------------ROAR
\begin{figure}[t]
% \fbox{\rule{0pt}{2in} \rule{0.9\linewidth}{0pt}}
\centering  
\begin{subfigure}[h]{0.5\linewidth}
     \includegraphics[width=\linewidth]{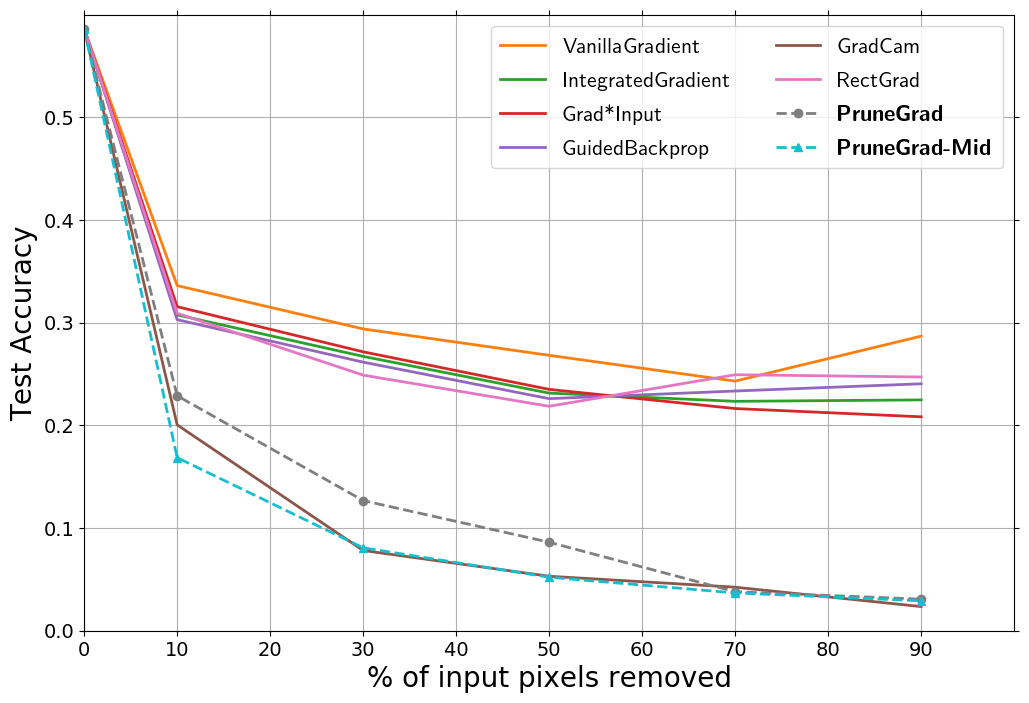}
     \caption{Birdsnap}
     \label{fig:roar_birdsnap}
\end{subfigure}\begin{subfigure}[h]{0.5\linewidth}
     \includegraphics[width=\linewidth]{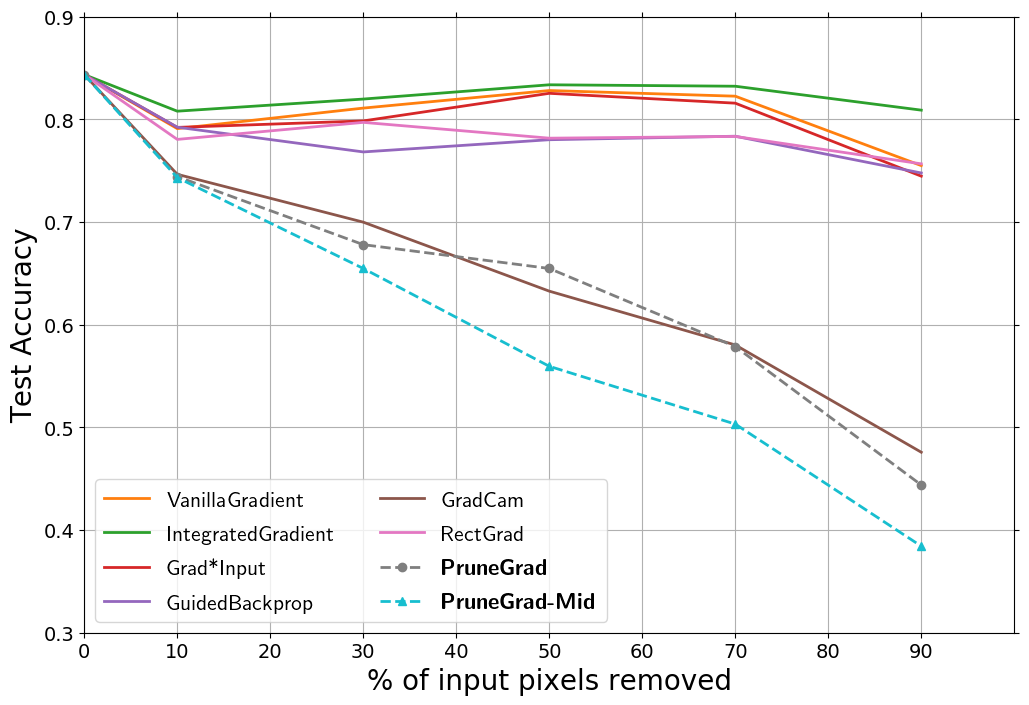}
     \caption{CIFAR-10}
     \label{fig:roar_cifar}
\end{subfigure}

\begin{subfigure}[h]{0.8\linewidth}
     \includegraphics[width=\linewidth]{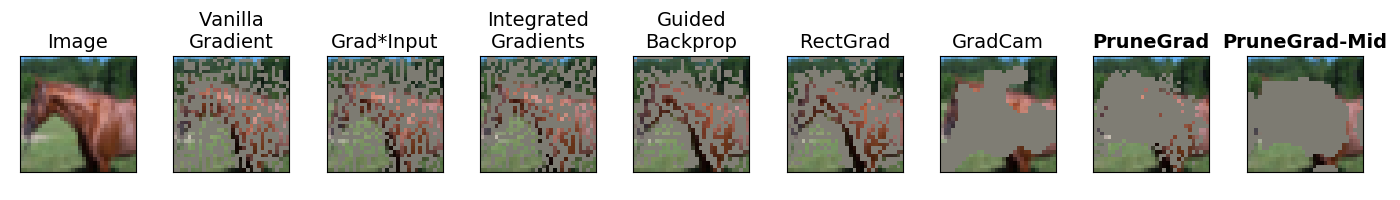}
     \caption{A sample modified image (50\% top pixels removed)}
     \label{fig:roar_cifar_qual}
\end{subfigure}
 \caption{\textbf{ROAR benchmark:} (a,b) Top $\{10,30,50,70,90\}\%$ of important pixels of each image (as assigned by each method) are replaced with a constant value. The drop in accuracy of the model after being retrained on the modified dataset signifies the effectiveness of each attribution method. Lower is better.}
%  of important pixels are modified (occluded) according to the importance score assignments of different attribution methods.
\label{fig:ROAR}
\end{figure}
%%%%%--------------------------
%% purpose and how it works
Hooker et al.~\cite{roar_neurips,hooker2018evaluating} recently proposed the Remove and Retrain (ROAR) benchmark for evaluating the importance of features for the prediction of the models.
Similar to Pixel Perturbation, ROAR is grounded on perturbing inputs based on attribution maps. 
However, pixel perturbation evaluation does not account for the fact that the change in output might be as a result of the network not having seen such perturbations during training. 
Therefore the ROAR benchmark retrains the network on the modified images.
The more the resulting accuracy on test set drops compared to the original network, the better does the feature attribution method highlight important features. 
%The experiment is performed for different percentiles of perturbed pixels.

%%%% Our Roar setting
We perform the experiments with top $\{10, 30, 50, 70, 90\}\%$ of pixels perturbed. 
The model is retrained for each attribution method (8 methods) on each percentile (5 percentiles) 3 times.
Due to the large number of retraining sessions required, we cannot report this benchmark on ImageNet. 
Similar to recent work (\cite{srinivas2019full}, custom small VGG network), we evaluate this benchmark on CIFAR-10 (60k images, 32$\times$32) with a ResNet-8 (three residual blocks). % using a custom small. 
Additionally, we evaluate ROAR using the Bridsnap dataset (45,653 images, 500 classes, cropped 224$\times$224) with a ResNet-50.

%%%% Explain Roar chart results
Fig.~\ref{fig:ROAR} presents the ROAR benchmark results. 
In order to better analyze these charts, we provide visual evidence from the modified images. 
In Fig. \ref{fig:roar_cifar_qual}, we present samples of modified images where the top 50\% of the pixels are removed according to the scores provided by different attribution methods. 
 
The maps of VanillaGradient, Input$\times$Gradient, and Integrated Gradients do not conceal the main object features in the image. 
After retraining, the model can still recognize the images. 
This phenomenon is also reflected in Fig.~\ref{fig:ROAR} and the results correlate with the reported results of \cite{roar_neurips}. 
It is noteworthy that Integrated Gradients complies with axioms such as completeness and sensitivity, however, the method is not performing well in highlighting highly contributing features, indicating the necessity for benchmarking in interpretability methods.

RectGrad and GuidedBackProp mostly highlight low-level features (e.g. edges) and the corresponding perturbed images are also still recognizable. 
As RectGrad is a strict variant of GuidedBackProp they are expected to highlight similar features and achieve similar results in the ROAR benchmark, which is visible both in the images in \ref{fig:roar_cifar_qual}, the charts (Fig.~\ref{fig:roar_birdsnap} and Fig.~\ref{fig:roar_cifar}). 

Fig.~\ref{fig:roar_cifar_qual} shows that the modifications resulting from GradCAM, PruneGrad and PruneGrad-Mid fully occlude the main features in these images, and this is also reflected in Fig.~\ref{fig:roar_birdsnap} and Fig.~\ref{fig:roar_cifar} where GradCAM, PruneGard and PruneGrad-mid significantly outperform other methods in highlighting important features in both experiments. 
However, although both PruneGrad and PruneGrad-Mid provide more fine-grained maps compared to GradCAM, only PruneGrad-Mid outperforms GradCAM. 
We postulate that this is due to interpolation step that exists in both GradCAM and PruneGrad-Mid. 
As the grainy structure of the PruneGrad maps does not entirely cover the features, the model still has access to parts of features during the retraining.
The smoothness of GradCAM and PruneGrad-Mid entirely covers the features due to interpolation.

%---------------------------------------------
\subsection{Attributions for Intermediate Layers}\label{exp:intermediate_layer}
Sec.~\ref{method:pruneGradMid} extended PruneGrad to find attributions of \emph{hidden} neurons. 
In order to visualize the high dimensional $\mathbb{R}^{k_{l}\times h_{l} \times w_{l}}$ neuron attributions of layer $l$, we sum the absolute values of all gradients across channel dimension $k_l$. 
The absolute value shows both the negatively contributing and positively contributing neurons. We refer to this method as PruneGrad-Mid and in our quantitative experiments the output of second to last residual block is used. 
The results are visualized for different layers of ResNet-50 in Fig.~\ref{fig:intermediate_layers}.
With decreasing resolution of deeper layers, the attribution map becomes less detailed. 
However, the method still attributes the correct part of the image. 
%---------------------------------------------

\subsection{Perturbations for Pruned Networks}\label{exp:adversarial}
In Sec.~\ref{method:reasoning}, we show the connection between pruning and input perturbation. 
Here, we perform two experiments that support this analysis. 
%-------------------------
\begin{figure}[t]

\begin{minipage}{.5\textwidth}
  \centering
  \includegraphics[width=.9\linewidth]{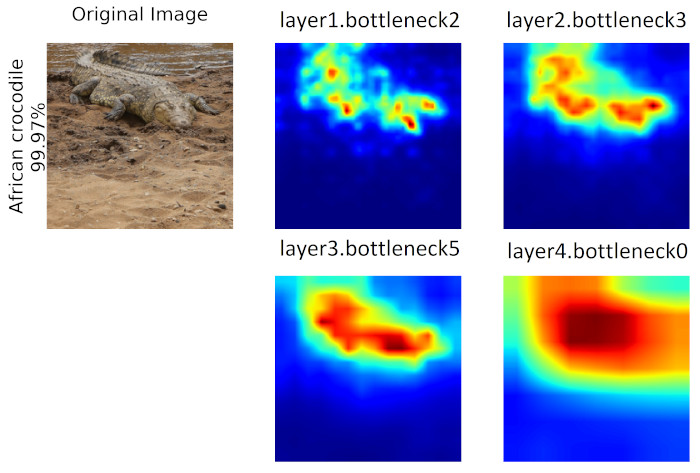}
  \captionof{figure}{PruneGrad-Mid on different layers of ResNet-50.}
  \label{fig:intermediate_layers}
\end{minipage}% 
\hspace{0.15cm}
\begin{minipage}{.5\textwidth}
  \centering
  \includegraphics[width=1\linewidth]{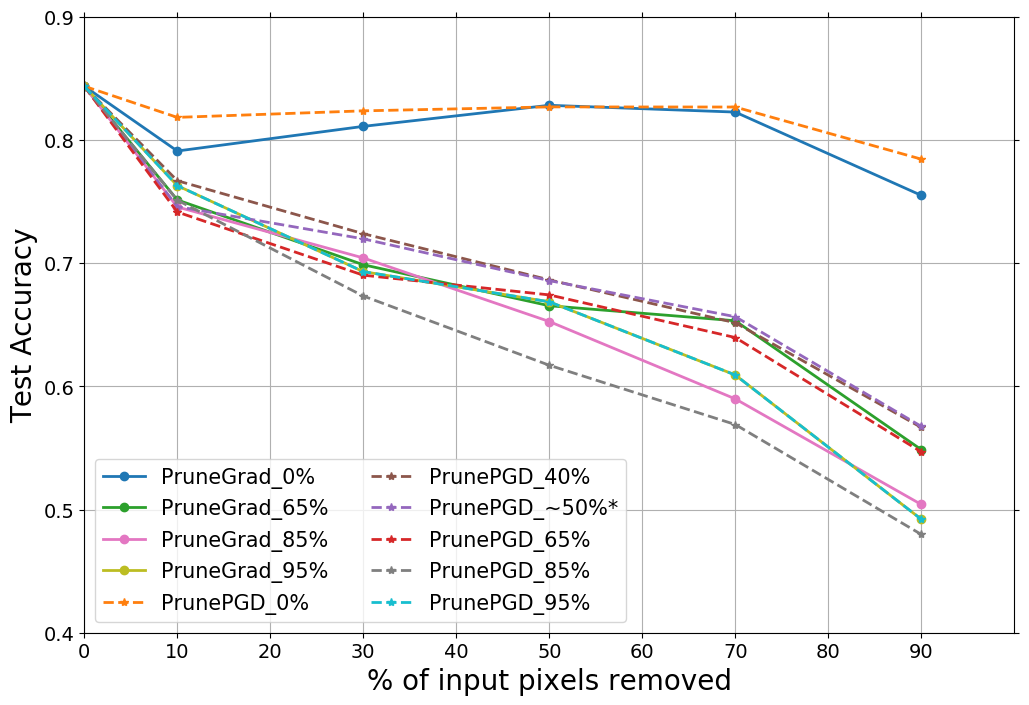}
  \captionof{figure}{ROAR evaluation for gradient and PGD perturbation on the pruned network for different pruning thresholds.}
  \label{fig:prune_grad_pgd_diff_thresholds}
\end{minipage}
\end{figure}
%----------------------

\textbf{Gradient as a Surrogate for Perturbation.} 
As explained earlier, PruneGrad is an approximate solution to the perturbation of the pruned network (Eq.~\ref{eq:maximization}). 
We provide experimental evidence using the ROAR framework to show that the approximate solution is sufficiently close to more accurate solutions of Eq.~\ref{eq:maximization}. 
We use projected gradient descent (PGD) which is the strongest first order method~\cite{madry2017towards} for solving Eq.~\ref{eq:maximization}. 
We compare pruning and gradient vs. pruning and perturbation (PrunePGD) on ROAR metric. 
PGD is applied using following parameters: $20$ iterations, step size $0.01$ and an L2 bound of $0.1$.
In Fig.~\ref{fig:prune_grad_pgd_diff_thresholds} PruneGrad and PrunePGD (with same pruning level) achieve approximately equivalent scores in ROAR. 
In addition, results of PrunePGD in sanity checks and pixel perturbation have been respectively reported in Fig.~\ref{fig:sanity} and \ref{fig:Perturb}. 
In all benchmarks, PruneGrad and PrunePGD perform similarly, justifying the proposed approximation which comes at one order of magnitude lower computational cost.
\textbf{Adversarial Perturbations on Input-Specific Pruned Networks.}
Adversarial perturbations are values for $\delta$ in Eq.~\ref{eq:maximization}, that are pathological and not human interpretable~\cite{goodfellow2014explaining,fong2017interpretable}. 
The term ‘human-interpretable’ casts subjectiveness over the properties of these features. 
For the first time, we use a quantitative method --- ROAR --- for a better understanding of adversarial perturbations, by quantitatively measuring whether adversarial perturbations are in fact highly contributing features. 
Fig.~\ref{fig:prune_grad_pgd_diff_thresholds} shows that for the original unpruned model (PrunePGD\_0), adversarial perturbations modify unimportant features for the model, as its ROAR curve shows the lowest drop in accuracy. 
When pruning unimportant neurons and therefore removing paths that activate features that are not already in the image, the adversarial perturbations reflect more important features that the model uses in the dataset. This is confirmed in Fig.~\ref{fig:prune_grad_pgd_diff_thresholds}; as pruning increases, perturbations reflect more important features.

One interesting observation is the difference of ROAR curves between perturbations for the $0\%$ pruned network and for a network where only dead neurons (zero activation) are removed, which is referred to as PrunePGD\_50* in Fig.~\ref{fig:prune_grad_pgd_diff_thresholds} (note that on average $50\%$ of neurons are dead in ResNet-8 on CIFAR-10).
This is equivalent to adversarial defenese of Wagner et al.~\cite{wagner2019interpretable} where during perturbation, gradients are only propagated if a neuron was active in the original image. 
Note that at this level, the method is reflecting contributing features (as opposed to $0\%$ pruning), quantitatively confirming the adversarial defense of \cite{wagner2019interpretable}.

%
%
%%%%---------------------
%
%-------------------------------------------------------------------------
\section{Conclusion}
In this work, we proposed input-specific pruning for improving feature attribution. 
We showed that by input-specific pruning, gradients reflect global (instead of local) importance information. 
The proposed attribution method is extensively evaluated using quantitative benchmarks: sanity checks, pixel perturbation, and ROAR.
Overall, we consistently outperform other methods. We provided a theoretical justification for PruneGrad using adversarial perturbations by showing that the gradient is a surrogate for adversarial perturbations on the pruned network.
We experimentally validated the explanation and introduced evaluating the importance of adversarial perturbations using ROAR feature evaluation.

\subsubsection{Acknowledgement}
The project is partly financially supported by Siemens Healthineers. Christian Rupprecht is supported by ERC IDIU-638009.
%
%
%
% ---- Bibliography ----
%
% BibTeX users should specify bibliography style 'splncs04'.
% References will then be sorted and formatted in the correct style.
%
\bibliographystyle{splncs04}
\bibliography{Refs}
\end{document}